\documentclass[10pt,twocolumn,letterpaper]{article}

\usepackage{cvpr}              
\usepackage[accsupp]{axessibility}  

\usepackage{graphicx}
\usepackage{times}
\usepackage{helvet}
\usepackage{courier}
\usepackage{amsmath}
\usepackage{algorithm}
\usepackage{algorithmic}
\usepackage{csquotes} 
\usepackage{color}
\usepackage{paralist}
\usepackage{amssymb}
\usepackage{indentfirst}
\usepackage{float}
\usepackage{multirow}
\usepackage{cite}
\usepackage{mathrsfs}

\usepackage{mathrsfs}
\usepackage[colorlinks, linkcolor=red,  anchorcolor=blue, citecolor=blue]{hyperref}
\usepackage{textcomp,booktabs}
\usepackage{amssymb}
\usepackage{pifont}
\usepackage{colortbl}
\definecolor{mygray}{gray}{.9}
\definecolor{mypink}{rgb}{.99,.91,.95}
\definecolor{mycyan}{cmyk}{.3,0,0,0}
\usepackage[capitalize]{cleveref}
\crefname{section}{Sec.}{Secs.}
\Crefname{section}{Section}{Sections}
\Crefname{table}{Table}{Tables}
\crefname{table}{Tab.}{Tabs.}

\def\cvprPaperID{1} 
\def\confName{NFVLR}
\def\confYear{2023}

\begin{document}

\title{ Learning CLIP Guided Visual-Text Fusion Transformer for Video-based Pedestrian Attribute Recognition }

\author{
Jun Zhu$^{1}$$^{\dag}$, Jiandong Jin$^{2}$$^{\dag}$, Zihan Yang$^{1}$, Xiaohao Wu$^{1}$, Xiao Wang$^{1}$\thanks{$^{\dag}$ denotes equal contribution. Corresponding author: Xiao Wang, email: xiaowang@ahu.edu.cn} \\ 
$^{1}$ School of Computer Science and Technology, Anhui University, Hefei 230601, China \\  
$^{2}$ School of Artificial Intelligence, Anhui University, Hefei 230601, China
}


\maketitle

\begin{abstract}
Existing pedestrian attribute recognition (PAR) algorithms are mainly developed based on a static image. However, the performance is not reliable for images with challenging factors, such as heavy occlusion, motion blur, etc. In this work, we propose to understand human attributes using video frames that can make full use of temporal information. Specifically, we formulate the video-based PAR as a vision-language fusion problem and adopt pre-trained big models CLIP to extract the feature embeddings of given video frames. To better utilize the semantic information, we take the attribute list as another input and transform the attribute words/phrase into the corresponding sentence via split, expand, and prompt. Then, the text encoder of CLIP is utilized for language embedding. The averaged visual tokens and text tokens are concatenated and fed into a fusion Transformer for multi-modal interactive learning. The enhanced tokens will be fed into a classification head for pedestrian attribute prediction. Extensive experiments on a large-scale video-based PAR dataset fully validated the effectiveness of our proposed framework. Both the source code and pre-trained models will be released at \url{https://github.com/Event-AHU/VTF_PAR}.  
\end{abstract}

\section{Introduction} 

Pedestrian Attribute Recognition (PAR)~\cite{wang2022PARsurvey, cheng2022VTB} is a very important research topic in computer vision and gets boosted greatly with the help of deep learning. Many representative PAR models are proposed in recent years based on convolutional neural networks (CNN)~\cite{he2016resnet}, and recurrent neural networks (RNN)~\cite{chung2014gru}. Wang et al.~\cite{wang2017RNNPAR} propose the JRL which learns the attribute context and correlation in a joint recurrent learning manner using LSTM~\cite{hochreiter1997lstm}. The self-attention based Transformer networks are first proposed to handle the natural language processing tasks and then are borrowed into the computer vision community~\cite{vaswani2017Transformer, DosovitskiyViT, wang2023MMPTMsSurvey, zhao2023transformerVLT, wang2021TNL2K} due to their great performance. Some researchers also exploit the Transformer for the PAR problem to model the global context information~\cite{tang2022drformer, cheng2022VTB}. DRFormer~\cite{tang2022drformer} is proposed to capture the long-range relations of regions and relations of attributes. VTB~\cite{cheng2022VTB} is also developed to fuse the image and language information for more accurate attribute recognition. In addition to understanding the pedestrian images using the attributes, this task also serves other computer vision problems, such as object detection~\cite{zhang2020attributedetection}, person re-identification~\cite{zheng2022progCMreid}, etc. Despite the great success of PAR, these works are developed based on a single RGB frame only which ignores the temporal information and maybe obtains sub-optimal results in practical scenarios.

As mentioned in work~\cite{chen2019videoPAR}, the video frames can provide more comprehensive visual information for the specific attribute, but the static image fails to. The authors propose to understand human attributes using video clips and propose large-scale datasets for video-based PAR. They also build a baseline by proposing the multi-task video-based PAR framework based on CNN and temporal attention. Better performance can be obtained on their benchmark datasets, however, we think the following issues still limit their overall results. 
\textbf{Firstly}, they adopt CNN as the backbone network to extract the feature representation of input images which learns the local features well. As is known to all, global relation in the pixel-level space is also very important for fine-grained attribute recognition. Several researchers resort to the Transformer network to capture such global information~\cite{DosovitskiyViT, vaswani2017Transformer}, however, their models can work for image-based attribute recognition only. 
\textbf{Secondly}, the authors formulate the video-based PAR as a multi-task classification problem and try to learn a mapping from a given video to attributes. The attribute labels are transformed into binary vectors for network optimization. However, the high-level semantic information is greatly lost which is very important for pedestrian attribute recognition.

\begin{figure} 
\center
\includegraphics[width=3.5in]{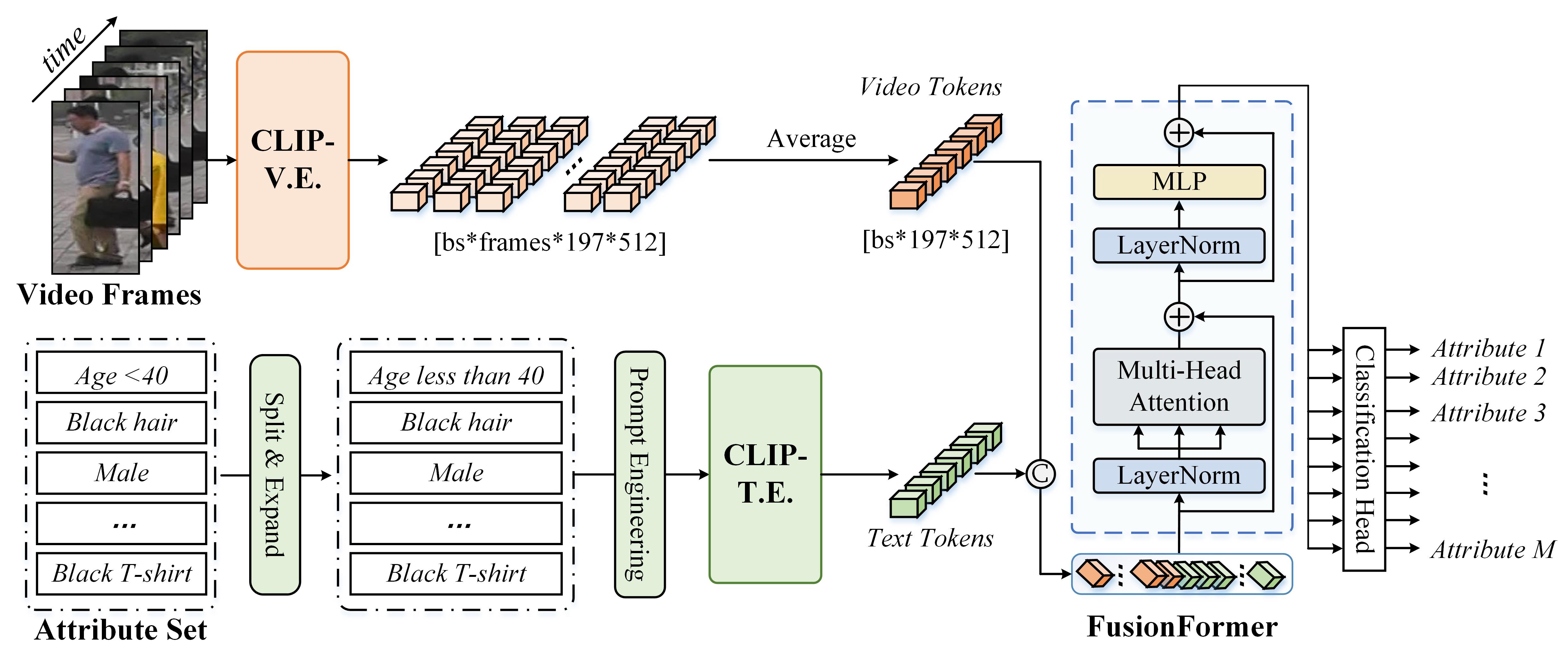}
\caption{ An overview of our proposed CLIP-guided Visual-Text Fusion Transformer for video-based PAR.} 
\label{framework}
\end{figure}

To address the aforementioned two issues, in this paper, we propose a novel CLIP-guided Visual-Text Fusion Transformer for Video-based Pedestrian Attribute Recognition. As shown in Fig.~\ref{framework}, we take the video frames and attribute set as the input and formulate the video-based PAR as a multi-modal fusion problem. To be specific, the video frames are transformed into video tokens using a pre-trained CLIP~\cite{radford2021CLIP} which is a multimodal big model. The attribute set is transformed into corresponding language descriptions using split, expand, and prompt engineering. Then, the text encoder of CLIP is used for the language embedding. After that, we concatenate the video and text tokens and feed them into a fusion Transformer for multi-modal information interaction which mainly contains layer normalization, multi-head attention, and MLP (Multi-Layer Perceptron). The output will be fed into a classification head for pedestrian attribute recognition.

To sum up, the main contributions of this paper can be concluded as following two aspects: 

$\bullet$ We propose a novel CLIP-guided Visual-Text Fusion Transformer for Video-based Pedestrian Attribute Recognition, which is the first work to address the video-based PAR from the perspective of visual-text fusion. 

$\bullet$ We introduce the pre-trained big model CLIP as our backbone network, which makes our model robust to the aforementioned challenging factors. Extensive experiments validated the effectiveness of our proposed model.

\section{Methodology} 

In this section, we will introduce our proposed framework from the following aspects, i.e., Input Processing and Embedding, Multimodal Fusion Transformer, and Optimization. 

\subsection{Input Processing and Embedding} 
Given the video frames $V = \{v_1, v_2, ..., v_T\}$ and attribute list $A = \{a_1, a_2, ..., a_M\}$, we first process these inputs to better utilize the pre-trained CLIP model. More in detail, the input frames are padded with zero pixels into a resolution $224 \times 224$, as the initial frames are slim but the pre-trained CLIP takes the fixed square resolution as the input. The ViT-B/16-based CLIP model is selected as the backbone considering its efficiency and accuracy. Therefore, the input frames are embedded into a set of visual tokens $T \times 197 \times 512$, here, 197 is the number of tokens and 512 is the dimension of each token. Then, we average this feature into a tensor $F_v \in \mathbb{R}^{197 \times 512} = \{f^1, f^2, ..., f^T\}$  along the temporal channel.  

For the attribute set $A$, we first need to process them into corresponding language descriptions to make full of CLIP text encoder. Specifically, we first split and expand each attribute to get the natural phrases. 
For example, ``\emph{Age $\leq$ 40}'' is processed into ``\emph{age less than 40}''. Then, prompt engineering is adopted to further process the phrases into natural language descriptions using carefully designed prompt templates. For example, \emph{age less than 40} is transformed into ``\emph{the pedestrian has an attribute \underline{age less than 40}}''. After all the attributes are processed, we adopt the text encoder of CLIP to get the text tokens $F_t = \{t^1, t^2, ..., t^M\}$. Then, the video and text tokens are concatenated together $[F_v, F_t]$ as the input of fusion Transformer.

\subsection{Multimodal Fusion Transformer} 
The Transformer network has been widely validated in its effectiveness in many research domains, including computer vision, natural language processing, and multimodal fusion~\cite{wang2023MMPTMsSurvey}. In this work, we adopt the Transformer to fuse the vision and language information to enhance the multimodal feature representation learning. Given the multimodal input $[F_v, F_t]$, we first adopt a layer normalization layer to process the input. Then, the input is transformed into three branches, i.e., $Q$, $K$, and $V$. We adopt the multi-head self-attention to learn the long-range global relations, and the basic operation of each self-attention layer can be described as $MLP(\mathrm{SAttn}({Q}, {K}) {V})$, where the $\mathrm{SAttn}$ is 
$\mathrm{SAttn}(Q, K) =  Softmax(\frac{{Q} {K}^{Tr}}{\sqrt{c}})$, 
where $c$ denotes feature dimension, $Tr$ is the transpose operation. 
The output tokens will be fed into a classification head for attribute prediction. Because the MARS dataset has 43 attributes, in our practical implementation, 43 fully connected layers are used as the classification head.

\subsection{Optimization} 
In this work, the video-based PAR task is formulated as a video-text fusion problem. Given the annotated attribute and raw video, we can train our framework in an end-to-end manner using supervised learning. The binary cross-entropy loss function is adopted for the optimization.

\section{Experiments}  

\subsection{Dataset, Metric, and Implementation Details} 
In our experiments, the MARS dataset~\cite{chen2019videoPAR} proposed by Chen et al. is used for both training and testing. The training subset of MARS contains 8,298 tracklets from 625 people, and the testing subset contains 8,062 tracklets corresponding to 626 pedestrians. For each tracklet, there are 60 frames on average. For the evaluation of our and the compared PAR models, we adopt the widely used Precision, Recall, and F1-score as the evaluation metric. Note that, the results reported in our experiments are obtained by averaging these metrics for multiple attribute groups. 

The ViT-B/16 version of pre-trained CLIP is used in our experiments. In the training phase, the parameters of CLIP encoder are fixed. The learning rate of our model is 0.001, weight decay is 1e-4. Our model is trained for a total of 20 epochs. The Adam~\cite{kingma2014adam} is adopted as our optimizer. Our model is implemented using Python and PyTorch~\cite{paszke2019pytorch} framework and trained on a server with RTX3090s.

\subsection{Compare with Other SOTA Models} 
In the experiments, we compare our model with multiple strong baseline methods on the MARS dataset, including 3DCNN~\cite{ji20123d}, CNN-RNN~\cite{mclaughlin2016recurrent}, VideoPAR~\cite{chen2019videoPAR}, and VTB~\cite{cheng2022VTB}. As shown in Table~\ref{resultsMARS}, we can find that our model beats all these compared methods by a large margin. Specifically, the VTB~\cite{cheng2022VTB} achieves $78.96, 78.42, 78.32$ on the Precision, Recall, and F1-score on this dataset, meanwhile, ours are $81.76, 82.95, 81.94$, the improvements are $+2.80, +4.53, +3.62$ on these metrics. Our results are also better than the VideoPAR proposed by Chen et al. (the video-based version, F1 score 72.04) by exceeding $+9.9$. For the fine-grained attribute results, we report them in Table~\ref{F1scoreMARS}. 
These experiments fully validated the effectiveness and advantages of our model.

\begin{table} 
\center
\scriptsize  
\caption{Results on MARS video-based PAR dataset. w/o denotes without the following module.}  
\label{resultsMARS} 
\begin{tabular}{l|c|ccc}
\hline \toprule [0.5 pt] 
\multicolumn{1}{c|}{\multirow{2}{*}{Methods}} & \multicolumn{1}{c|}{\multirow{2}{*}{Backbone}} & \multicolumn{3}{c}{MARS} \\ \cline{3-5} 
\multicolumn{1}{c|}{} &
\multicolumn{1}{c|}{} &
\multicolumn{1}{c}{Prec} &
\multicolumn{1}{c}{Recall} &
\multicolumn{1}{c}{F1}  \\ 
\hline
3DCNN~\cite{ji20123d}  & - & - & - & 61.87 \\
CNN-RNN~\cite{mclaughlin2016recurrent} & - & - & - & 70.42 \\
VideoPAR (Image)~\cite{chen2019videoPAR} & ResNet50 & - & - & 67.28 \\
VideoPAR (Video)~\cite{chen2019videoPAR} & ResNet50 & - & - & 72.04 \\
\hline 
VTB~\cite{cheng2022VTB} & ViT-B/16 & \textcolor{blue}{\textbf{78.96}} & \textcolor{blue}{\textbf{78.42}} & \textcolor{blue}{\textbf{78.32}}  \\
Ours                    & ViT-B/16 & \textcolor{red}{\textbf{81.76}} & \textcolor{red}{\textbf{82.95}} & \textcolor{red}{\textbf{81.94}} \\
Improvements            & - & \textcolor{green}{\textbf{+2.80}} & \textcolor{green}{\textbf{+4.53}} & \textcolor{green}{\textbf{+3.62}}          \\
\hline 
Ours (w/o FusionFormer)                 & ViT-B/16 & 77.60  & 81.32 &  78.69  \\
\hline \toprule [0.5 pt] 
\end{tabular} 
\end{table}

\begin{table} 
\center
\scriptsize  
\caption{Results on MARS video-based PAR dataset. F1-score are reported for all the assessed attributes.}   
\label{F1scoreMARS} 
\begin{tabular}{l|c|c|c|c|c}
\hline \toprule [0.5 pt] 
\multicolumn{1}{c|}{\multirow{2}{*}{Attribute}} & \multicolumn{1}{c|}{\multirow{1}{*}{VideoPAR}} & \multicolumn{1}{c|}{3DCNN}  & \multicolumn{1}{c|}{CNN-RNN}  & \multicolumn{1}{c|}{\multirow{1}{*}{VideoPAR}}  & \multicolumn{1}{c}{Ours} \\ 
\multicolumn{1}{c|}{} &
\multicolumn{1}{c|}{(Image)} &
\multicolumn{1}{c|}{} &
\multicolumn{1}{c|}{} &
\multicolumn{1}{c|}{(Video)} &
\multicolumn{1}{c}{}  \\ 
\hline
top length & 58.72 & 56.37 & 65.18 & \textcolor{blue}{\textbf{ 71.61}} & \textcolor{red}{\textbf{97.26}} \\
bottom length & 92.29 & 89.35 & 93.33 & \textcolor{red}{\textbf{93.90}} & \textcolor{blue}{\textbf{ 93.69}}\\
shoulder bag & 72.57 & 61.30 & \textcolor{blue}{\textbf{ 75.89}} & \textcolor{red}{\textbf{76.08}} & 65.61\\
backpack & 85.95 & 76.58 & \textcolor{blue}{\textbf{ 87.17}} & \textcolor{red}{\textbf{87.62}} & 82.08\\
hat & 57.57 & 57.69 & \textcolor{blue}{\textbf{ 77.74}} & \textcolor{red}{\textbf{77.84}} & 72.76\\
hand bag & 62.82 & 59.90 & \textcolor{blue}{\textbf{ 71.68}} & \textcolor{red}{\textbf{73.55}} & 59.08\\
hair & \textcolor{blue}{\textbf{ 86.91}} & 82.77 & 87.11 &\textcolor{red}{\textbf{88.17}} & 86.37\\
gender & 90.89 & 85.75 & 92.44 & \textcolor{blue}{\textbf{ 92.50}} & \textcolor{red}{\textbf{92.88}}\\
bottom type & 81 .69 & 72.86 & 84.16 & \textcolor{blue}{\textbf{ 86.62}} & \textcolor{red}{\textbf{97.21}}\\
pose & 56.91 & 47.69 & 58.36 & \textcolor{blue}{\textbf{ 61.36}} & \textcolor{red}{\textbf{74.84}}\\
motion & 39.39 & 33.64 & \textcolor{blue}{\textbf{ 43.92}} & 43.69 & \textcolor{red}{\textbf{93.50}}\\
top color & \textcolor{blue}{\textbf{ 72.72}} & 65.63 & 69.28 & 71.44 & \textcolor{red}{\textbf{74.97}}\\
bottom color & \textcolor{blue}{\textbf{ 44.63}} & 40.39 & 39.68 & 43.98 & \textcolor{red}{\textbf{69.76}}\\
age & 38.87 & 36.22 & 39.93 & \textcolor{blue}{\textbf{ 40.21}} & \textcolor{red}{\textbf{87.07}}\\
\hline
Average-F1 & 67.28 & 61.87 & 70.42 & \textcolor{blue}{\textbf{ 72.04}} & \textcolor{red}{\textbf{81.94}} \\
\hline \toprule [0.5 pt] 
\end{tabular} 
\end{table}

\subsection{Ablation Study} 

\noindent 
\textbf{Component Analysis. } 
In our proposed framework, the \emph{fusion Transformer} and \emph{pre-trained CLIP backbone} are our key components. In this section, we analyze the two components and report the recognition results in Table~\ref{resultsMARS}. 
The VTB~\cite{cheng2022VTB} is our baseline which adopts the standard ViT-B/16 model as the backbone, and it achieves 78.96/78.42/78.32 on Precision, Recall, and F1-score. When the CLIP model is used, the results can be improved to 81.76/82.95/81.94, which validated the effectiveness of the pre-trained big model for video-based PAR. 
When replacing the FusionFormer using regular fully connected layers, the results are dropped from $81.76, 82.95, 81.94$ to $77.60, 81.32, 78.69$, which demonstrates that this fusion module also contributes to our final performance.

\begin{figure} 
\center
\includegraphics[width=3in]{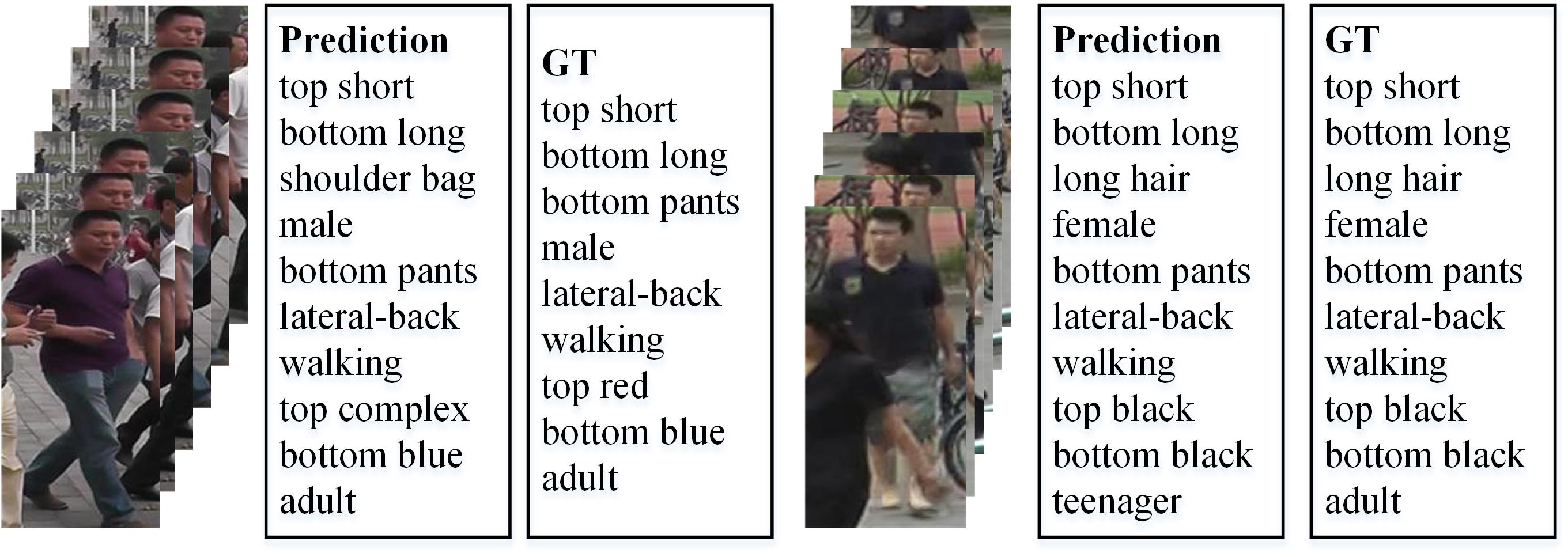}
\caption{Visualization of our predictions and Ground Truth (GT).} 
\label{visResults}
\end{figure}

\noindent 
\textbf{Analysis on Input Video Frames. }
The number of video frames plays an important role in video-based pedestrian attribute recognition. In this part, we train and test different numbers of input frames on the MARS dataset and report the experimental results in Table~\ref{frameAnalysis}. We can find that the performance can be gradually improved with the increase of video frames, i.e., the F1-score is ranging from 77.27 to 81.94. 

\begin{table}
\center
\small  
\caption{Results with different input frames on MARS dataset.} 
\label{frameAnalysis} 
\begin{tabular}{c|ccccc}
\hline \toprule [0.5 pt]
\multicolumn{1}{c|}{\multirow{1}{*}{\# Frames}}  &
  \multicolumn{1}{c}{6} &
  \multicolumn{1}{c}{4} &
  \multicolumn{1}{c}{2} & 
  \multicolumn{1}{c}{1} \\ 
  \hline
  Precision & 81.76  & 81.23 & 79.56 & 76.43  \\
  Recall & 82.95  & 82.32 & 81.20  & 78.88\\
  F1-score & 81.94 & 81.39 & 79.97 & 77.27  \\
    \hline \toprule [0.5 pt] 
\end{tabular} 
\end{table}

\noindent 
\textbf{Visualization.}  
In addition to the aforementioned quantitative analysis, we also give a qualitative analysis in this subsection. As shown in Fig.~\ref{visResults}, we can find that our model predicts human attributes accurately.

\section{Conclusion}  
Different from the mainstream image-based PAR, in this paper, we propose a novel CLIP-guided Visual-Text Fusion Transformer for video-based pedestrian attribute recognition, which makes full use of temporal information. More in detail, we formulate the video-based PAR as a vision-language fusion problem and adopt pre-trained big models CLIP to extract the feature embeddings of given video frames. To better utilize the semantic information, we take the attribute list as another input and fuse it with video tokens using a fusion Transformer. The enhanced tokens will be fed into a classification head for pedestrian attribute prediction. We conduct extensive experiments on a large-scale video-based PAR dataset and demonstrate that our model obtains superior recognition performance. In our future works, we will consider designing fine-grained partial region mining modules to realize a higher performance attribute recognition. Also, new prompt learning/tuning techniques are worthy to be exploited for pre-trained multi-modal big model guided video-based pedestrian attribute recognition.

\textbf{Acknowledgement} This paper is supported by the National Natural Science Foundation of China NO. 62102205.

{\small
\bibliographystyle{ieee_fullname}
\bibliography{reference}
}

\end{document}